\documentclass[5p,times]{elsarticle}

\usepackage{lineno,hyperref}
\modulolinenumbers[5]

\journal{Future Generation Computer Systems}

\usepackage[inline]{enumitem}
\usepackage{lipsum}
\usepackage{amsmath}
\usepackage{amsfonts}
\usepackage[table]{xcolor}
\usepackage{adjustbox}
\usepackage{graphicx}
\usepackage{tabularx}
\usepackage{verbatimbox}
\usepackage{amsthm}
\usepackage{booktabs}
\usepackage{multirow}
\usepackage{float}
\usepackage{tikz}
\usepackage{comment}
\usetikzlibrary{matrix}
\usetikzlibrary{decorations.text}
\usetikzlibrary{positioning}
\usepackage{url}
\usepackage{footnote}
\usepackage{microtype}
\setlist[enumerate]{label=\arabic*}
\usepackage{todonotes}
\usepackage{mathtools}
\usepackage{algorithmic}
\usepackage{algorithm}
\newtheorem{definition}{Definition}[section]

\newcommand{\correcciones}[1]{{\color{black}#1}}    
    
\newcommand{\nuria}[1]{{\color{black}#1}}    
    
\newcommand{\eugenio}[1]{{\color{black}#1}}
\newcommand{\paco}[1]{{\color{black}#1}}
\newcommand{\nrb}[1]{{\color{black}#1}}
\newcommand{\corr}[1]{{\color{black}#1}}
\newcommand{\emc}[1]{{\color{black}#1}}

\begin{document}

\begin{frontmatter}

\title{Dynamic Defense Against Byzantine Poisoning Attacks in Federated Learning}

\author[decsai]{Nuria~Rodr\'{i}guez-Barroso\corref{mycorrespondingauthor}}
\cortext[mycorrespondingauthor]{Corresponding author}
\ead{rbnuria@ugr.es}

\author[decsai]{Eugenio Mart\'{i}nez-C\'{a}mara}
\ead{emcamara@decsai.ugr.es}

\author[lsi]{M. Victoria Luz\'{o}n}
\ead{luzon@ugr.es}

\author[decsai]{Francisco Herrera}
\ead{herrera@decsai.ugr.es}

\address[decsai]{Department of Computer Science and Artificial Intelligence, AndalusianResearch Institute in Data Science and Computational Intelligence (DaSCI), University of Granada, 18071 Granada, Spain}
\address[lsi]{Department of Software Engineering, Andalusian Research Institute inData Science and Computational Intelligence (DaSCI), University of Granada, 18071 Granada, Spain}

\begin{abstract}
Federated learning, as a distributed learning that conducts the training on the local devices without accessing to the training data, is vulnerable to \nrb{Byzatine} poisoning adversarial attacks. We \emc{argue} that the federated learning model has to avoid those kind of adversarial attacks through filtering out the \emc{adversarial} clients \corr{by means of the federated aggregation operator}. We propose \emc{a} dynamic federated \corr{aggregation operator} that dynamically \emc{discards} those adversarial clients \emc{and} allows to prevent the corruption of the global learning model. We \emc{assess} it as a \nrb{defense against} adversarial \nrb{attacks} deploying a deep learning classification model in a federated learning setting \nrb{on} the Fed-EMNIST Digits, Fashion MNIST \nrb{and CIFAR-10} image datasets. The results show that the dynamic selection of the clients to aggregate enhances the performance of the global learning model and discards the adversarial and poor (with low quality models) clients.

\end{abstract}


\end{frontmatter}


\section{Introduction}
The standard machine learning approach is built upon an algorithm that learns from a centralized data source. Distributed machine learning proposes the distribution of the data and elements of a learning model among several \correcciones{nodes} as a solution for the unceasing growing of learning model complexity and the size of training data \cite{dean2012,chenxin2017}. However, the distributed machine learning solution is neither valid for the data privacy challenge, nor for an scenario with a large number of clients and a non homogeneous data distribution \cite{MAL-083, CHEN20211}.



Federated learning (FL) is a machine learning approach in which the algorithms \nuria{learn} from sequestered data \cite{MAL-083,MOTHUKURI2021619}. \emc{The FL model} is mainly composed of two components: a global server that owns the global learning model and a set of clients storing the local learning models and the local training datasets. Likewise, FL consists in: \begin{enumerate*}[label={(\arabic*)}] \item training the local learning models in each data source, \item distilling the parameters of the local learning models into a central server, \item aggregating the parameters of the local models \corr{in the global learning model} and \item updating the local learning models with the aggregated federated \corr{global} \emc{learning} model after the aggregation\end{enumerate*}. This specific setting supports its main feature, which is the prevention of data leakage and the protection of data privacy, \emc{since} the data do not abandon its local storage and they are not shared with any other client \nuria{or third party}. \corr{Since FL  is  a  user  privacy-preserving approach designed to decentralized scenarios, an Artificial Intelligence of Things (AIoT) setting is a natural way to use it, for both the distributed nature and the privacy needed in IoT (Internet of Things) devices \cite{9302596}}.

Machine learning is vulnerable to malicious manipulations on the input data or the learning model to cause incorrect classification \cite{dalvi2004}. This vulnerability becomes harder to address in FL due to most of the defensive approaches are based data inspection techniques. Among the different kind of adversarial attacks in the literature \cite{huang2011}, in this paper we focus on \corr{byzantine} poisoning attacks \cite{Biggio2014}, which are based on the arbitrary manipulation of the training data (data poisoning attack \cite{gu2019,jagielski2018}) or the client model updates (model poisoning attacks \cite{bhagoji2018model}) \corr{with the aim of hindering the performance of the FL model}.

We \emc{argue} in this paper that the FL model has to be able to \eugenio{dynamically} avoid adversarial clients to preserve the learning model from \corr{byzantines} \eugenio{poisoning attacks}, which is usually performed on the server by the federated aggregation operator. In the literature there are a number of federated aggregation operators, but they do not prevent the federated model from this kind of attacks \cite{red,bib:konecny16federatedoptimizacion,wang2017}, or they do it following some assumptions about the nature of the adversarial clients \cite{fung2018} \nrb{or prove to be insufficiently effective \cite{pmlr-v108-bagdasaryan20a}}.

\corr{We propose \emc{the Dynamic Defense Against Byzantine Attacks (DDaBA), which} is a dynamic aggregation operator that dynamically selects the clients to be aggregated and discards those ones considered as adversarial, and it features agnostic about the number and nature of the adversarial clients. This dynamic defense is built upon an Induced Ordered Weighted Averaging (IOWA) operator \cite{iowa_yager_filev}, \emc{which  aggregates the clients on a weighted basis according to an induced-ordered function and a linguistic quantifier.} We use as induced-ordered function the performance of the local learning models on a validation set stored in the server. \emc{The linguistic quantifier addresses the weighting of the clients, which usually depends on the knowledge of the problem and predefined parameters. We design an agnostic linguistic quantifier on the nature of the problem, which is based on:} \begin{enumerate*}[label={(\arabic*)}]\item considering the distribution of data resulting from measuring the performance variation between local learning models on the validation set, \item assuming that the resulting distribution follows an exponential distribution, and \item \emc{using the properties of that distribution to set the parameters of the linguistic quantifier in order to discard the adversarial clients that correspond to outliers in the exponential distribution according to the Tukey criteria.}\end{enumerate*}}

\corr{We evaluate \emc{the} DDaBA as a defense in a FL model for image classification. For that purpose, w}e leverage the benchmark image classification datasets Fed-EMNIST\footnote{\url{htt ps://www.nist.gov/node/1298471/emnist-dataset}} Digits \cite{emnist}, Fashion MNIST\footnote{\url{https://github.com/zalandoresearch/fashion-mnist}} \cite{DBLP:journals/corr/abs-1708-07747} and \nrb{CIFAR-10,\footnote{\url{https://www.cs.toronto.edu/~kriz/cifar.html}}} and we distribute the data over the clients following a non independent and identically distributed (non-IID) distribution. We compare \emc{the DDaBA} with the classical federated aggregation operator \corr{FedAvg \cite{red} with no defense} and the state-of-the-art defenses against three different byzantine attacks: label-flipping \cite{DBLP:journals/corr/abs-2007-08432}, out-of-distribution \cite{fort2021exploring} and random weights  \cite{DBLP:journals/corr/abs-1911-12560} attacks. We show that the \nrb{DDaBA} is able to identify the adversarial \paco{and poor} clients, filter them out and enhance the performance of the global learning model. 


\emc{We analyze the behavior of the DDaBA in an scenario with a extreme proportion of adversarial clients, and we see that the performance of the federated \corr{global} model is hindered. Although this is} a very unlikely scenario, we \emc{also} introduce \emc{the} static version \emc{of DDaBA}, Static Defense Against Byzantine Attacks (SDaBA), which \emc{predefine the parameters of the linguistic quantifier of the IOWA operator for discarding the susceptible adversarial clients.} \emc{The SDaBA, as well as the DDaBA, outperforms all the baselines in the three adversarial attacks developed for the evaluation.}

The rest of the work is organized as follows: the following section summarizes the background related to FL, adversarial attacks in FL and defenses against them. Section \ref{s_proposal} is focused on the description of the dynamic FL model for identifying adversarial clients. \nuria{We detail the experimental set-up in Section \ref{s_experimental_study} and evaluate and analyze the results of the FL models in Section \ref{s_results}. Finally, conclusions are described in Section \ref{s_conclusions}.} 

\section{Background}\label{ss_rw}

We \emc{expound} in this section some relevant concepts and related works. We \corr{introduce} FL in Section \ref{s_fl}, we \nrb{\corr{describe} the main types of adversarial attacks in FL} in Section \ref{s_adv_att}, and \nrb{we detail the proposed defenses against byzantine attacks} in Section \ref{ss_def}.

\vspace{1cm}

\subsection{Federated Learning}\label{s_fl}

\eugenio{FL is a learning approach pushed by the need of overcoming the limitations of distributed learning for preserving data privacy and for processing large number of clients following a non homogeneous data distribution \cite{konecny2016}. FL proposes a new training approach of learning algorithms that consists in the iterative training of the model in the devices that own the data, the aggregation of those models in the federated model, and the updating of the local models with the federated model. \emc{Hence}, FL prevents from data leakage and preserves data privacy, \emc{since} the data do not leave the electronic device.}

Formally, FL is a distributed machine learning paradigm consisting of a set of clients $\{C_1, \dots, C_n\}$ with their respective local training data $\{D_1, \dots, D_n\}$. Each of these clients $C_i$ has a local learning model named as $L_i$ represented by the parameters $\{L_1, \dots, L_n\}$. FL aims at learning \emc{a} global learning model represented by $G$, using the scattered data across clients through an iterative learning process known as \textit{round of learning}. For that purpose, in each round of learning $t$, each client trains its local learning model over their local training data $D^t_i$\emc{,} which updates the local parameters $L^{t}_i$ to $\hat{L}^t_i$. Subsequently, the global parameters $G^t$ are computed aggregating the trained local parameters $\{\hat{L}^t_1, \dots, \hat{L}^t_n\}$ using an specific federated aggregation operator $\Delta$\emc{,} and the local learning models are updated with the aggregated parameters:

\begin{equation}
\begin{split}
    G^t = \Delta(\hat{L}^t_1,\hat{L}^t_2, \dots, \hat{L}^t_n)\\
    L^{t+1}_i \leftarrow G^t, \quad \forall i \in \{1, \dots, n\}
\end{split}
    \label{eq_fl_aggregation}
\end{equation}

The updates among the clients and the server are repeated as much as needed for the learning process. Thus, the final value of $G$ will sum up the knowledge sequestered in the clients.

\subsection{Related works about adversarial attacks}\label{s_adv_att}

\nrb{Machine learning is highly susceptible to adversarial attacks \cite{laskov2010}, and the vast majority of the defensive approaches are based on three \emc{approaches} \cite{huang2011}: \begin{enumerate*}[label=(\arabic*)]
\item game theory \cite{nilesh2004}, \item data sanitation \cite{nelson2009} and \item resilient and robust learning models, which assume that a fraction of the training data may be manipulated and consider it as outliers~\cite{croux2007}.\end{enumerate*} Due to the federated aggregation operator is agnostic in relation with adversarial clients information, the first approach can not be applied in FL. Likewise, since the training data in FL is inaccessible by the server, the second approach is also not feasible in FL. Therefore, the most promising defense approach is developing resilient and robust federated aggregation operators with the ability to safeguard the model from the effect of attacks.

According to \cite{DBLP:journals/corr/abs-2012-06337}, there are two types of adversarial attacks in FL:  \begin{enumerate*}[label=(\arabic*)]\item \textit{Inference attacks} \cite{Nasr_2019}, which aim at inferring information from the training data; and \item \textit{poisoning attacks} \cite{8975792}, which pursue to compromise the global learning model. \end{enumerate*} Concerning inference attacks, there are different types of them depending on the information being inferred. The most important ones are the property and membership inference attacks, which respectively seek to infer certain properties of the data and the membership of specific samples in the training set. Because of their nature, the defenses proposed in the literature are based on applications derived from or inspired by the Differential Privacy \cite{TCS-042}. Regarding poisoning attacks, we identify two taxonomies: 
\begin{enumerate}
\item Depending on which part of the FL model is attacked, we differentiate between \textit{model-poisoning} \cite{fi13030073} and \textit{data-poisoning attacks} \cite{DBLP:journals/corr/abs-2004-10020}. In practice, both are almost equivalent, since a poisoning of the data results in a poisoned model. However, data-poisoning attacks and some of the model-poisoning attacks fail to be effective since the attack dissipates in the aggregation of many clients. Hence, these attacks are combined with \textit{model-replacement} \cite{pmlr-v108-bagdasaryan20a} techniques, which boosts the adversarial model (or models) in order to replace the global model.
\item Depending on the purpose of the attack, we distinguish between \textit{untargeted or byzantine attacks} \cite{10.1145/3335772.3335936}, which seek to affect the model's performance, and \textit{targeted or backdoor attacks} \cite{pmlr-v108-bagdasaryan20a}, which aim at injecting a secondary or backdoor task into the global model by stealth. 
\end{enumerate}}

\subsection{Defenses against adversarial attacks} \label{ss_def}

The literature provides multiple solutions to both byzantine and backdoor attacks in classical machine learning. The vast majority of these defenses are based on data inspection methods, such as removing outliers from the training data in centralized learning \cite{10.5555/3294996.3295110} or, in a distributed setting, removing outliers from participant's training data or models \cite{DBLP:journals/corr/abs-1811-09904, Shen2016AurorDA}. In both cases, the available defenses require data inspection, which is not possible in FL. Therefore, defenses against adversarial attacks in FL must be designed ad hoc.

Regarding the state-of-the-art defenses designed to be applied in federated settings, they are based on the modification of the aggregation operator, because the attack is usually carried out by the clients. The most important defenses against byzantine attacks are based on a more robust aggregation of the updates and they are called \textit{byzantine-robust aggregation rules}. We highlight the following ones:
\begin{itemize}
    \item Coordinate-wise aggregations \cite{pmlr-v80-yin18a}, which replaces the mean of the classical aggregation operator FedAvg \cite{red} with more robust statistics to outliers or anomalous data. The \emc{main ones} are the trimmed-mean and the median. 
    \item Krum (and MultiKrum) \cite{NIPS2017_f4b9ec30}, which is based on using geometric properties to determine the most central model updates vectors. This defense requires a $k$ hyper-parameter \emc{that} determines the number of clients remaining in the aggregation.
    \item Bulyan \cite{pmlr-v80-mhamdi18a} which \emc{is} the state of the art. It is built as a combination of Krum and trimmed-mean. Accordingly, the model updates vectors are sorted according to their geometrical centrality and are aggregated through a trimmed-mean with a $m$ parameter, which discards a total of $2m$ clients.
\end{itemize}

Additionally, differential privacy \cite{TCS-042} \emc{methods} are an important safeguard for the information shared during the communication between the server and the clients. Therefore, the defensive challenges of the FL should focus on client attacks.

The main weakness of the defenses proposed in the literature is that they are highly dependent on parameters, which beforehand are difficult to set without information about the number or nature of the adversary clients. Thus, we propose in this paper a defense mechanism against poisoning attacks, which dynamically selects the clients that are not adversarial and filters out the adversarial or the poor ones (clients with low quality models) \nrb{without the requirement to set any parameters}.

\section{\nrb{Dynamic Defense Against Poisoning Attacks}} \label{s_proposal}

FL is featured by its restriction to access to the training data, which is sequestered in the clients. Accordingly, poisoning attacks, \nrb{both data and (local) model poisoning} \cite{gu2019,jagielski2018}, grounded in the malicious manipulation of the training data \nrb{or the local model updates}, can corrupt the FL model, which cannot inspect the training to defend itself against this kind of adversarial attacks.

We propose \nrb{a defense against byzantine poisoning attacks built upon a federated aggregation operator based on a Induced Ordered Weighted Averaging (IOWA) \cite{iowa_yager_filev} that dynamically selects the clients to be aggregated, and filters out the adversarial ones. We call it Dynamic Defense Against 
Byzantine Attacks (DDaBA).}

The IOWA operators, and more generally the Ordered Weigh\-ted Averaging (OWA) ones \cite{owa_1}, are functions for weighting the contribution of a set of clients in a aggregation process, as it is the aggregation of the parameters of the local learning models in FL. We mathematically \paco{introduce} OWA and IOWA operators in Appendix \ref{appendix:iowa}, and according to the definition the IOWA operator is composed of \begin{enumerate*}[label=(\arabic*)]\item an order-inducing function to set the weighting assignation order, and \item a linguistic quantifier to calculate the weight contribution value.\end{enumerate*} We define the induced-order function used in DDaBA in Section \ref{s_fl_iowa_fla}, and the linguistic quantifier that dynamically adapts the weighting value calculation during the FL training in Section \ref{ss_lq}. \nuria{Finally, we sum up DDaBA in Section \ref{ss_ddaba}.}

\subsection{Accuracy-based induced ordering function for \nrb{clients model updates}}
\label{s_fl_iowa_fla}

\eugenio{The aim of \nrb{byzantine} poisoning adversarial attacks is hindering the performance of a FL model through altering the training data or \nrb{directly the model updates}. Since FL is grounded in the aggregation of the $L_{i}$, those maliciously altered ones would perform lower than the non-altered ones. Hence, the validation of the $L_{i}$ before the aggregation may help to identify the suspicious adversarial clients.}

We propose the Local Accuracy Function\eugenio{\paco{,} $f_{LA}$\paco{,}} to measure the performance of each $L_i$ \eugenio{before its aggregation}. \eugenio{The $f_{LA}$ function is based on the availability of a validation set shared among the clients. The viability of this validation set is justified by its reduced size compared to the size required for training, and the possibility of making it up through expert or prior knowledge. We define the $f_{LA}$ function in Definition \ref{def_laf}.}

\begin{definition}[Local Accuracy Function ($f_{LA}$)]\label{def_laf}
    it measures the performance of a local learning model $L_i$ using \nuria{a fixed} validation dataset \correcciones{named as $VD$}. For that, it computes the \textit{accuracy} of $L_i$ over $VD$:
    
    \begin{equation}
        \correcciones{f_{LA}(L_i) = \text{accuracy}(L_i, VD)}
    \end{equation}
\paco{where $\text{accuracy}(L_i, VD)$ refers to the standard accuracy evaluation measure of the local learning model $L_i$ in the dataset $VD$.}
\end{definition}

\nrb{Once the clients model updates are sorted according to this sorting function, we expect that the benign client's models will converge to a common solution, while the adversarial client's models will not, but \emc{they} will converge to a worse solution for the original problem. Therefore, if we define the random variable resulting from the differences in accuracy among all clients with the client \emc{that} scored the highest accuracy as follows:}

\nrb{
\begin{equation}
    \mathbb{X}_i^{f_{LA}} = \max_{i}\{f_{LA}(L_i)\} - f_{LA}(L_i).
\end{equation}}

\nrb{We assume that this random variable $\mathbb{X}$ will approximate an Exponential Distribution, since there will be many values close to zero (and always positive), and very few far from zero. }

\subsection{Dynamic linguistic quantifier for weighting \nrb{the contribution of clients}}\label{ss_lq}

The non-IID data distribution of most of the FL settings make impossible to know beforehand the nature of the clients, and hence it is impossible to know the amount of adversarial clients. Therefore, the selection of the FL clients by its weighted contribution has to be dynamically calculated for adapting to the nature of the clients.

The dynamic selection of the \nrb{DDaBA} model is based on a IOWA linguistic quantifier that some of its parameters values depend on \nrb{the resulting exponential distribution after ordering the clients model updates $\mathbb{X}_i^{f_{LA}}$}. Before the definition of the linguistic quantifier of \nrb{DDaBA}, we first define the IOWA linguistic quantifier in Definition \ref{df_lq_paper}.

\begin{definition}[Linguistic quantifier]\label{df_lq_paper}
It is a function $Q:[0,1] \rightarrow [0,1]$ verifying $Q(0) = 0$, $Q(1) = 1$ and $Q(x) \ge Q(y)$ for $x > y$. Equation \ref{eq_q_weight_calculation_1_paper} defines how the function $Q$ computes the weighting values where $w_i$ represents the weighting associated to the position $i$ of a vector of dimension $n$, and Equation \ref{eq_standard_quantifier_1_paper} defines the behaviour of the function $Q$. 

\begin{equation}
    w_i^{(a,b)} = Q_{a,b} \left( \frac{i}{n} \right) - Q_{a,b} \left( \frac{i-1}{n} \right)
    \label{eq_q_weight_calculation_1_paper}
\end{equation}

\begin{equation}\label{eq_standard_quantifier_1_paper}
    Q_{a,b}(x) = 
    \begin{dcases}
        0 & 0\leq x \leq a\\
        \frac{x-a}{b-a} & a \leq x \leq b \\
        1 & b \leq x \leq 1
    \end{dcases}
\end{equation}
\eugenio{where $a, b \in [0,1]$ satisfying $0 \leq a \leq b \leq 1$, and they set the intervals for calculating the contribution weight of each $L_{i}$. For the sake of clarification, those $x$ values in the same interval will have the same weighting value.}
\end{definition}

\begin{equation}\label{eq_dynamic_quantifier}
    Q_{a,b,c,y_b}(x) = 
    \begin{dcases}
        0 & 0\leq x \leq a\\
        \frac{x-a}{b-a} \cdot  y_b & a \leq x \leq b \\
        \frac{x-b}{c-b} \cdot (1-y_b) + y_b & b \leq x \leq c \\
        1 & c \leq x \leq 1
    \end{dcases}
\end{equation}

We redefine the function $Q_{a,b}$ for providing it a dynamic behaviour and a higher weighting of top clients, which depends on \nrb{the random variable $\mathbb{X}_i^{f_{LA}}$}. Accordingly, we propose $Q_{a,b,c,y_b}$ that is defined in Equations \ref{eq_dynamic_quantifier}, and incorporates two new parameters to the model \nrb{(c and $y_b$), in addition to the two existing ones. The definition of each of the parameters is as follows:}

\nrb{
\begin{enumerate} 
\item Parameter $a$. This parameter represents the proportion of clients to which null weighing is assigned. Since we do not want to filter out those clients which stand out "at the top", i.e. those that obtain the best accuracy, we set the value to 0.
\item Parameter $b$. It sets the portion of clients we consider as top clients and we want to weight higher. The choice of this parameter is done dynamically, so that the top clients correspond to the first decile of the distribution of  $\mathbb{X}_i^{f_{LA}}$. Formally, $b$ is the portion of clients that verify
\begin{equation}
    \mathbb{X}_i^{f_{LA}} \leq \frac{\ln(10/9)}{\lambda},
\end{equation}
where $\lambda = \frac{1}{\mu_{\mathbb{X}_i^{f_{LA}}}}$ and $\mu_{\mathbb{X}_i^{f_{LA}}}$ the mean of $\mathbb{X}_i^{f_{LA}}$.

\item The dynamic behavior of the parameter $c$. This parameter represents the portion of clients that we do not discard. For example, a value of $c = 0.8$ means that the 20\% of the clients will be discarded. With the aim of dynamically adapt it in each aggregation, we identify the problem of filtering out adversarial clients as a problem of outlier detection in $\mathbb{X}_i^{f_{LA}}$. We thus employ the Tukey criteria \cite{tukey77, tukey} for anomalies in exponential probability distribution functions and set $c= 1 - \hat{c}$ where $\hat{c}$ is the portion of clients that verify

\begin{equation}
    \mathbb{X}_i^{f_{LA}} \geq Q_3 + 1.5 IQR = \frac{\ln(4)}{\lambda} + 1.5 \frac{\ln(3)}{\lambda},
\end{equation}

where $\lambda = \frac{1}{\mu_{\mathbb{X}_i^{f_{LA}}}}$ and $\mu_{\mathbb{X}_i^{f_{LA}}}$ the mean of $\mathbb{X}_i^{f_{LA}}$.
\item Parameter $y_{b}$. It provides the weighting of the top clients together with $b$. In particular, it represents the portion of the total weight assigned to these clients. In order to weight the top clients with double the importance of the rest of the clients participating in the aggregation, we set
\begin{equation}
    y_b = \frac{2 |Top|}{2 |Top| - |Rest|},
\end{equation}
where $|Top| = b \times n$ and $|Rest| = (c-b) \times n$.
\end{enumerate}}


\nrb{Analogously to Equation \ref{eq_q_weight_calculation_1_paper}, we obtain the weighting of each client from the $Q_{a,b,c,y_b}$ function according to Equation \label{q_nueva}.}


\begin{equation}
    w_i^{(a,b,c,y_b)} = Q_{a,b,c,y_b} \left( \frac{i}{n} \right) - Q_{a,b,c,y_b} \left( \frac{i-1}{n} \right)
    \label{q_nueva}
\end{equation}

\nrb{\subsection{Defense based on the federated aggregation}\label{ss_ddaba}}

\nrb{Finally, using the equations defined above and the definitions of FL (Equation \ref{eq_fl_aggregation}), we define DDaBA as a defense \corr{consisting of} the following aggregation operator:

\begin{equation}
    \corr{DDaBA(\{\hat{L}^t_1, \hat{L}^t_2, \dots, \hat{L}^t_n\}, VD) = \sum_{i=1}^n w_i^{(a,b,c,y_b)} \hat{L}^t_i}
\end{equation}

where $w_i^{(a,b,c,y_b)}$ is defined in Equation \ref{q_nueva} and $\hat{L_i^t}$ the local model update of the client $i$ for $i \in \{1, \dots, n\}$}. \corr{Algorithm \ref{alg:pseudocode} depicts the DDaBA pseudo-code.}

\setlength{\textfloatsep}{10pt}
\begin{algorithm}[h!]
   \caption{DDaBA}
   \label{alg:pseudocode}
\begin{algorithmic}
   \STATE {\bfseries Input:} local updates $\{\hat{L}^t_1, \hat{L}^t_2, \dots, \hat{L}^t_n\}$ and $VD$
     
    \STATE Initialize $G^{t}$
    
    \FOR {$i=0$ {\bfseries to} $n$}
        \STATE $f_{LA}(L_i) = \text{accuracy}(L_i, VD)$
    \ENDFOR
    
    \FOR {$i=0$ {\bfseries to} $n$}
        \STATE  $\mathbb{X}_i^{f_{LA}} = \max_{i}\{f_{LA}(L_i)\} - f_{LA}(L_i)$
    \ENDFOR

    \STATE $a = 0$
    \STATE $b = | \mathbb{X}_i^{f_{LA}} \leq \frac{\ln(10/9)}{\lambda} |$
    \STATE $c = |\mathbb{X}_i^{f_{LA}} \geq \frac{\ln(4)}{\lambda} + 1.5 \frac{\ln(3)}{\lambda}| $
    \STATE $y_b = \frac{2 |b \times n|}{2 |b \times n| - |(c-b) \times n|}$
    
   \FOR {$i=0$ {\bfseries to} $n$}
        \STATE $w_i = w_i^{(a,b,c,y_b)}$ according to Equation \ref{q_nueva}.
    \ENDFOR
    
   \STATE $G_{t} = \sum_{i=0}^n w_i \hat{L}_i^t$
   
   \STATE \textbf{Return}  $G_{t}$
   
\end{algorithmic}
\end{algorithm}

\section{Experimental set-up} \label{s_experimental_study}

The evaluation of DDaBA is performed \corr{by means of the accuracy of the resulting FL model} in three datasets arranged for FL, and we describe them in Section \ref{dataset}. Also, we deployed an image classification deep learning model in the FL setting. \corr{Since the main aim of this work is to propose a dynamic defense against byzantine attacks, we use an standard CNN-based image classification model composed of two CNN layers followed by its corresponding max-pooling layers, a dense layer and the output layer with a softmax activation function for the Fed-EMNIST and Fashion MNIST datasets and a pre-tained model based on EfficientNet \cite{DBLP:journals/corr/abs-1905-11946} for the CIFAR-10 dataset.} Finally, the federated aggregation operators used as baselines are introduced in Section \ref{s_baselines} and the covered attacks in Section \ref{s_scenarios}.

\subsection{Evaluation datasets}\label{dataset}

Since the \nrb{DDaBA} needs a validation set for dynamically discarding adversarial clients, we create it from the test subsets of the \nrb{three} datasets, by assigning 20\% of the sample in the test dataset to the validation set. The \nrb{three} datasets used in the evaluation are described as what follows:

\begin{enumerate}
    \item The Fed-EMNIST (Federated Extended Modified NIST) dataset, which was presented in 2017 in \cite{emnist} as an extension of the MNIST dataset \cite{lecun98}. The EMNIST Digits class contains a balanced subset of the digits dataset containing 28,000 samples of each digit. The dataset consists of 280,000 samples, which 240,000 are training samples and 40,000 test samples. We use its federated version by identifying each client with an original writer.

    \item The Fashion MNIST \cite{DBLP:journals/corr/abs-1708-07747} aims to be a more challenging replacement for the original MNSIT dataset. It contains a balanced subset of the 10 different classes containing 7,000 samples of each class. Hence, the dataset consists of 70,000 samples, which 60,000 are training samples and 10,000 test samples. We set the number of clients to 500.
    
    \item \nrb{The CIFAR-10 dataset is a labeled subset of the 80 million tiny images dataset \cite{4531741}. It consists of 60000 32x32 color images in 10 classes, with 6000 images per class. There are 50000 training images and 10000 test images, which correspond to 1000 images of each class. We set the number of clients to 100.}
    
\end{enumerate}

In summary, the datasets, after appropriate modifications to prepare the validation sets, follow the data distributions shown in Table \ref{tab:datasets}.

\begin{table}[!h]
\centering
\caption{Size of the training, validation and test sets of Fed-EMNIST, Fashion MNIST and CIFAR-10 datasets.}
\label{tab:datasets}
\resizebox{\linewidth}{!}{\begin{tabularx}{\linewidth}{@{}Xrrr@{}}
\toprule
 & \textbf{Training} & \textbf{Validation} & \textbf{Test}\\
\midrule
\textbf{Fed-EMNIST} & 240,000 & 8,000 & 32,000\\
\textbf{Fashion MNIST} & 60,000 & 2,000 & 8,000\\
\textbf{CIFAR-10} & 60,000 & 2,000 & 8,000\\
\bottomrule

\end{tabularx}}
\label{tab:4_nonad_emnist}
\end{table}

With the aim of adapting both Fashion MNIST and CIFAR-10 datasets to a federated environment, the training data is distributed among the clients following a non-IID distribution. Accordingly, we randomly assign instances of a reduced number of labels to each client simulating a scenario in which each client contains partial information.

\vspace{0.5cm}

\subsection{Baselines based on federated aggregation operators} \label{s_baselines}

\nrb{We compare the DDaBA defense with the classical federated aggregation operator FedAvg \cite{bib:mcmahan16communicationefficient} and the following state-of-the-art defenses against byzantine poisoning attacks:
\begin{itemize}
\item \textit{Median} \cite{chen2017distributed}. It is one of the byzantine-robust aggregation rules which is based on replacing the mean with the median in the aggregation method, which is more robust against extreme values.
\item \textit{Trimmed-mean} \cite{DBLP:journals/corr/abs-1803-01498}. It represents another byzantine-robust aggregation rule. It relies on using a more robust version of the mean that consists in eliminating a fixed percentage ($k$) of extreme values both below and above the data distribution.
\item \textit{Krum and Multikrum} \cite{NIPS2017_f4b9ec30}. It sorts the clients according to the geometric distances of their model updates distributions and chooses the one closest to the majority as the aggregated model. Multikrum incorporates an $d$ parameter\emc{,} which specifies the number of clients to be aggregated (the first $d$ after being sorted) resulting in the aggregated model. 
\item \textit{Bulyan} \cite{pmlr-v80-mhamdi18a}. It represents the state-of-the-art combining Krum and the thrimmed-mean. Hence, it sorts the clients according to their geometric distances and, according to an $f$ parameter, filters out the $2f$ clients of the tails of the sorted distribution of clients and aggregates the rest of them.
\end{itemize}

The main weakness of Multikrum and Bulyan is that they strongly depend on a parameter given by the user. Both \emc{are} optimal if the number of adversarial clients is known, which is usually not the case.

}

\subsection{Byzantine Data and Model Poisoning Attacks} \label{s_scenarios}

There are multitude of byzantine adversarial attacks both data and model poisoning. Due to the high number of clients participating in the aggregation and the low proportion of clients that will be adversarial in a reasonable configuration, poisoning attacks are very ineffective as their effect dissipates in the aggregation. For that reason, poisoning attacks are combined with model-replacement \cite{pmlr-v108-bagdasaryan20a} techniques, which weight the contribution of adversarial clients in the aggregation according to a boosting parameter that is distributed among the adversarial clients. 

\corr{The adversarial attacks covered in this work are the following:}
\begin{itemize}
    \item \textit{Label-flipping attack} \cite{DBLP:journals/corr/abs-2007-08432}. It is a data poisoning attack consisting of randomly flipping the labels of the adversarial attacks. This way, the adversarial clients learn incorrect information that send to the server.
    
    \item \textit{Out-of-distribution attack} \cite{fort2021exploring}. It is another data poisoning attack consisting of introducing into the adversarial clients' training dataset some samples out of the training distribution. In practice, the most frequent approaches are to introduce samples from another dataset with the same features (e.g. EMNIST and Fashion MNIST) or to introduce randomly generated samples. We adopt the second approach in the experimentation.
    
    \item \textit{Random weights} \cite{DBLP:journals/corr/abs-1911-12560}. It is a model poisoning attack based on randomly generate the model updates of each adversarial client.
\end{itemize}

\begin{table*}[ht!]
\caption{Mean results for the label-flipping byzantine attack in terms of accuracy. We also show, in the first row, the expected accuracy with \textit{FedAvg} but without any attack. The best result for each of the scenarios is highlighted in bold.}
\label{tab:labelflipping}
\begin{center}
\addvbuffer[5pt 5pt]{\resizebox{\textwidth}{!}{\begin{tabular}{l|rrr|rrr|rrr}
\toprule
 & \multicolumn{3}{c|}{\textbf{Federated EMNIST}} & \multicolumn{3}{c|}{\textbf{Fashion MNIST}} & \multicolumn{3}{c}{\textbf{CIFAR-10}}\\
\toprule
 & \textbf{1-out-of-30} & \textbf{5-out-of-30} & \textbf{10-out-of-50} & \textbf{1-out-of-30} & \textbf{5-out-of-30} & \textbf{10-out-of-50} & \textbf{1-out-of-30} & \textbf{5-out-of-30} & \textbf{10-out-of-50}\\
\rowcolor[HTML]{FFFFFF} 
\midrule
\textbf{No attack} & 0,9657 & 0,9657 & 0,9629 & 0,8719 & 0,8719 &  0,8697 & 0,8357 & 0,8357 & 0,8231\\
\midrule
\rowcolor[HTML]{FFFFFF} 
\textbf{FedAvg} & 0,1591 & 0,4212 & 0,4007 & 0,1917 & 0,3665 &  0,4322 & 0,1184 & 0,1436 & 0,2448 \\
\midrule
\rowcolor[HTML]{FFFFFF} 
\textbf{Trim.-mean} & 0,9428 & 0,8739 & 0,8370 & 0,8672 & 0,8325 &  0,861 & 0,8239 & 0,7346 & 0,8220 \\
\rowcolor[HTML]{FFFFFF} 
\textbf{Median} & 0,9313 & 0,9161 & 0,9097 & 0,8671 & 0,8473 & 0,8585 & 0,8287 & 0,8090 & 0,8289\\
\rowcolor[HTML]{FFFFFF} 
\midrule
\textbf{Krum} & 0,8917 & 0,8706 & 0,8634 & 0,7264 & 0,7197 & 0,7473 & 0,7479 & 0,7610 & 0,7698 \\
\rowcolor[HTML]{FFFFFF} 
\textbf{MultiKrum (5)} & 0,9132 & 0,9270 & 0,9189 & 0,8403 & 0,8433 &  0,8255 & 0,8164 & 0,8232 & 0,8114 \\
\rowcolor[HTML]{FFFFFF} 
\textbf{MultiKrum (20)} & 0,9563 & 0,9571 & 0,9504  & 0,8727 & 0,8724 &  0,8680 & 0,8439 & 0,8479 & 0,8518 \\
\rowcolor[HTML]{FFFFFF} 
\midrule
\textbf{Bulyan (f=1)} & 0,9523 & 0,7813 & 0,5809 & 0,8689 & 0,7830 &  0,7875 & 0,8265 & 0,6595 & 0,6454 \\
\rowcolor[HTML]{FFFFFF} 
\textbf{Bulyan (f=5)} & 0,9365 & 0,9421 & 0,9516  & 0,8617 & 0,8652 & 0,8726 & 0,8492 & 0,8451 & 0,8540  \\
\rowcolor[HTML]{FFFFFF} 
\midrule
\textbf{DDaBA} & \textbf{0,9657} & \textbf{0,9663} & \textbf{0,9643}  & \textbf{0,8817} & \textbf{0,8783} & \textbf{0,8807} & \textbf{0,8633} & \textbf{0,8503} & \textbf{0,8557}\\
\bottomrule
\end{tabular}}}
\end{center}
\end{table*}

\begin{table*}[ht!]
\caption{Mean results for the label-flipping byzantine attack in terms of accuracy. We also show, in the first row, the expected accuracy with \textit{FedAvg} but without any attack. The best result for each of the scenarios is highlighted in bold.}
\label{tab:ood}
\begin{center}
\addvbuffer[5pt 5pt]{\resizebox{\textwidth}{!}{\begin{tabular}{l|rrr|rrr|rrr}
\toprule
 & \multicolumn{3}{c|}{\textbf{Federated EMNIST}} & \multicolumn{3}{c|}{\textbf{Fashion MNIST}} & \multicolumn{3}{c}{\textbf{CIFAR-10}}\\
\toprule
 & \textbf{1-out-of-30} & \textbf{5-out-of-30} & \textbf{10-out-of-50} & \textbf{1-out-of-30} & \textbf{5-out-of-30} & \textbf{10-out-of-50} & \textbf{1-out-of-30} & \textbf{5-out-of-30} & \textbf{10-out-of-50}\\
\rowcolor[HTML]{FFFFFF} 
\midrule
\textbf{No attack} & 0,9657 & 0,9657 & 0,9629 & 0,8719 & 0,8719 & 0,8697  & 0,8357 & 0,8357 & 0,8231\\
\midrule
\rowcolor[HTML]{FFFFFF} 
\textbf{FedAvg} &0,4093 & 0,4404 & 0,4350 & 0,2041 & 0,3667 & 0,4657 & 0,1468 & 0,1922 & 0,3419 \\
\midrule
\rowcolor[HTML]{FFFFFF} 
\textbf{Trim.-mean} & 0,9456 & 0,8602 & 0,8531 & 0,8652 & 0,8348 & 0,8310 & 0,8202 & 0,7441 & 0,7400 \\
\rowcolor[HTML]{FFFFFF} 
\textbf{Median} & 0,9345 & 0,9200 & 0,9144 & 0,8662 & 0,8465 & 0,8454 & 0,8223 & 0,8019 & 0,8073\\
\rowcolor[HTML]{FFFFFF} 
\midrule
\textbf{Krum} & 0,8693 & 0,8668 & 0,8621 & 0,7361 & 0,7062 & 0,7281 & 0,7202 & 0,7310 & 0,7408 \\
\rowcolor[HTML]{FFFFFF} 
\textbf{MultiKrum (5)} & 0,9169 & 0,9330 & 0,9198 & 0,8493 & 0,8430 & 0,8345 & 0,8305 & 0,8191 & 0,8023 \\
\rowcolor[HTML]{FFFFFF} 
\textbf{MultiKrum (20)} & 0,9545 & 0,9544 & 0,9506 & 0,8747 & 0,8719 & 0,8733 & 0,8607 & 0,8519 & 0,8521 \\
\rowcolor[HTML]{FFFFFF} 
\midrule
\textbf{Bulyan (f=1)} & 0,9507 & 0,7872 & 0,5812 & 0,8704 & 0,7601 & 0,6930 & 0,8319 & 0,6862 & 0,5551 \\
\rowcolor[HTML]{FFFFFF} 
\textbf{Bulyan (f=5)} & 0,9353 & 0,9383 & 0,9502 & 0,8713 & 0,8654 & 0,8757 & 0,8440 & 0,8498 & 0,8481  \\
\rowcolor[HTML]{FFFFFF} 
\midrule
\textbf{DDaBA} & \textbf{0,9652} & \textbf{0,9620} & \textbf{0,9654} & \textbf{0,8761} & \textbf{0,8841} & \textbf{0,8783} & \textbf{0,8626} & \textbf{0,8599} & \textbf{0,8632}\\
\bottomrule
\end{tabular}}}
\end{center}
\end{table*}

\begin{table*}[ht!]
\caption{Mean results for the label-flipping byzantine attack in terms of accuracy. We also show, in the first row, the expected accuracy with \textit{FedAvg} but without any attack. The best result for each of the scenarios is highlighted in bold.}
\label{tab:random-weights}
\begin{center}
\addvbuffer[5pt 5pt]{\resizebox{\textwidth}{!}{\begin{tabular}{l|rrr|rrr|rrr}
\toprule
 & \multicolumn{3}{c|}{\textbf{Federated EMNIST}} & \multicolumn{3}{c|}{\textbf{Fashion MNIST}} & \multicolumn{3}{c}{\textbf{CIFAR-10}}\\
\toprule
 & \textbf{1-out-of-30} & \textbf{5-out-of-30} & \textbf{10-out-of-50} & \textbf{1-out-of-30} & \textbf{5-out-of-30} & \textbf{10-out-of-50} & \textbf{1-out-of-30} & \textbf{5-out-of-30} & \textbf{10-out-of-50}\\
\rowcolor[HTML]{FFFFFF} 
\midrule
\textbf{No attack} & 0,9657 & 0,9657 & 0,9629 & 0,8719 & 0,8719 & 0,8697  & 0,8357 & 0,8357 & 0,8231\\
\midrule
\rowcolor[HTML]{FFFFFF} 
\textbf{FedAvg} & 0,0997 & 0,0994 & 0,1001 & 0,1006 & 0,1016 & 0,0997 & 0,0998 & 0,0994 & 0,1005  \\
\midrule
\rowcolor[HTML]{FFFFFF} 
\textbf{Trim.-mean} & 0,9537 & 0,1039 & 0,0990 & 0,8751 & 0,1004 & 0,0999 & 0,8608 & 0,0992 & 0,0998  \\
\rowcolor[HTML]{FFFFFF} 
\textbf{Median} & 0,9367 & 0,9354 & 0,9342 & 0,8654 & 0,8618 & 0,8554 & 0,8499 & 0,8664 & 0,8646 \\
\rowcolor[HTML]{FFFFFF} 
\midrule
\textbf{Krum} & 0,8314 & 0,8652 & 0,8541 & 0,7156 & 0,7459 & 0,7342 & 0,7184 & 0,7164 & 0,7994 \\
\rowcolor[HTML]{FFFFFF} 
\textbf{MultiKrum (5)} & 0,9325 & 0,9228 & 0,9191 & 0,8348 & 0,8343 & 0,8278 & 0,8164 & 0,8115 & 0,8167  \\
\rowcolor[HTML]{FFFFFF} 
\textbf{MultiKrum (20)} & 0,9565 & 0,9577 & 0,9510 & 0,8764 & 0,8751 & 0,8676 & 0,8488 & 0,8488 & 0,8531 \\
\rowcolor[HTML]{FFFFFF} 
\midrule
\textbf{Bulyan (f=1)} & 0,9598 & 0,0997 & 0,0998 & 0,0990 & 0,1001 & 0,0990 & 0,8529 & 0,0996 & 0,0993  \\
\rowcolor[HTML]{FFFFFF} 
\textbf{Bulyan (f=5)} & 0,9379 & 0,9377 & 0,9514 & 0,8746 & 0,8690 & 0,8746 & 0,8502 & 0,8411 & 0,8519  \\
\rowcolor[HTML]{FFFFFF} 
\midrule
\textbf{DDaBA} & \textbf{0,9653} & \textbf{0,9645} & \textbf{0,9622} & \textbf{0,8801} & \textbf{0,8778} & \textbf{0,8777} & \textbf{0,8656} & \textbf{0,8624} & \textbf{0,8626}\\
\bottomrule
\end{tabular}}}
\end{center}
\end{table*}

We experiment with four different settings of adversarial clients for each of the previously described attacks:

\begin{itemize}
    \item \textit{1-out-of-30 attack scenario}. Consisting of 1 adversarial clients of a total of 30 clients participating in each aggregation, which represents $1/30$ of adversarial clients.
    \item \textit{5-out-of-30 attack scenario}. Consisting of 5 adversarial clients of a total of 30 clients participating in each aggregation, which represents $1/6$ of adversarial clients.
    \item \textit{10-out-of-50 attack scenario}. Consisting of 5 adversarial clients of a total of 50 clients participating in each aggregation, which represents $1/5$ of adversarial clients.
\end{itemize}

In each of the scenarios described, the boosting factor is divided by the number of adversarial clients in order to carry out the model-replacement.

\subsection{Implementation details}

\textcolor{black}{We provide the code of DDaBA federated aggregtion operator\footnote{\url{https://github.com/ari-dasci/S-DDaBA}} in order to ensure the reproducibility of the experiments.
Due to the large number of existing FL frameworks \cite{rodrguezbarroso2020federated} and with the aim of showing that DDaBA is independent of the framework, we have selected two of them: 
\begin{itemize}[noitemsep]
    \item The Sherpa.ai FL \cite{rodrguezbarroso2020federated} framework.
    \item The Flower \cite{beutel2020flower} framework. 
\end{itemize}
For each framework, we include Jupyter notebooks to visualise how the aggregation operator works and to facilitate its understanding.}

\section{Experimental results}\label{s_results}

We evaluate the performance of DDaBA as a defense to the byzantine attacks described in Section \ref{s_adv_att} in two ways: \begin{enumerate*}[label={(\arabic*)}] \item In Section \ref{s_analysis}, we compare the behavior of DDaBA \corr{in terms of the performance of the resulting FL model} with the baselines described in Section \ref{s_baselines} and, \item In Section \ref{static} \emc{we analyze DDaBD in a scenario with a high number of adversarial clients, and we propose a modification of it for this particular scenario.}\end{enumerate*}


\subsection{Analysis of the results}\label{s_analysis}

Tables \ref{tab:labelflipping}, \ref{tab:ood} and \ref{tab:random-weights} show the results obtained in label-flipping, out-of-distribution and random weights attacks. Regarding the strength of the attacks, we find that all three are sufficiently powerful to pose a challenge to defenses. \corr{In fact, notice that the attack is slightly more effective when there are fewer adversarial clients since the boosting factor is divided among fewer clients.} The out-of-distribution attack is slightly less damaging while the random weights attack achieves the lowest performance without defense, ranking as the most challenging. The results obtained both in the different types of attacks and in the considered datasets confirm common conclusions, so we discuss all the results as a whole.

When evaluating the performance of the baselines we hereby confirm that MultiKrum and Bulyan do indeed represent the state of the art. However, they are highly dependent of the $d$ and $f$ parameters since they set the number of clients to keep or discard, respectively, in the aggregation. For example, in the 10-out-of-50 scenario and Bulyan with $f=1$ we verify this weakness, since only $2f = 2$ clients would be discarded from the aggregation, which is not enough to defend the model in the presence of 10 adversarial clients. A possible solution would be to set this value always to high, but this is also a limitation because in the case of having fewer adversarial clients than $2f$ the quality of the model decreases (e.g., 1-out-of-30 using Bulyan with $f=5$). Finally, MultiKrum and Bulyan promise optimal performance in the case of knowing the number of adversarial clients, which is not the case. This enhances the need for a defense that dynamically estimates how many clients to filter in the aggregation.

In contrast, the outperformance of DDaBA is confirmed in all the attack settings considered enhancing its success regardless of the type of attack and the proportion of adversarial clients. Moreover, DDaBA achieves better results than the no attack situation in the vast majority of the scenarios. This is because the dynamic filtering of clients not only discards those that are adversarial but also those that perform too poorly to contribute to improving the global learning model.

\subsection{Extreme attack scenarios - A static version of DDaBA}\label{static}

It has been proven that discarding clients based on whether or not they are outliers in a distribution formed from performance on a common validation set overcomes the defenses of the state of the art. However, this approach based on data distributions has a weakness \nrb{stemmed from} the fact that the distribution we use to search outliers is configured with the same data that we subsequently evaluate. Therefore, with a very high presence of adversarial clients, the resulting distribution will be highly skewed by this data, resulting in no outlier. Although we recognize this weakness, we point out that it is not a major one\emc{,} since it is highly unlikely for the percentage of adversarial clients in an FL scenario to be so high as to cause the defense to fail.

To overcome this weakness, we propose a static version of DDaBA called Static Defense Against Byzantine Attacks (SDaBA), which incorporates the only difference that the proportion of clients to be discarded from the aggregation is computed using a fixed parameter $\alpha$. In particular, instead of eliminating those clients that represent outliers in the distribution $\mathbb{X}_i^{f_{LA}}$, we eliminate those clients whose distance to the best accuracy is greater than $\alpha$ times the maximum of the distances. In other words, using $\mathbb{X}_i^{f_{LA}}$, we set $c = 1 - \beta$ where $\beta$ is the portion of clients verifying that 

\begin{equation}
    \mathbb{X}_i^{f_{LA}} \geq \alpha \mathbb{X}_n^{f_{LA}} \quad \forall i \in \{1, \dots, n\}
\end{equation}

in Equations \ref{eq_dynamic_quantifier} and \ref{q_nueva}. Analogously, we set $b=0.2$ in order to consider as top clients the top 20\% clients.

With the aim of evaluating SDaBA we set $\alpha=1/4$ and the 10-out-of-30 attack scenario consisting of 10 adversarial clients of a total of 30 clients participating in each aggregation, which represents $1/3$ of adversarial clients, which is an unusual high proportion of them. Table \ref{tab:extreme-scenario} shows the results of DDaBA and SDaBA in comparison with the baselines in this extreme attack scenario in Federated EMNIST.

\begin{table}[ht!]
\caption{Mean results for the extreme scenario (10-out-of-30) in Federated EMNIST in terms of accuracy. We also show, in the first row, the expected accuracy with \textit{FedAvg} but without any attack. The best result for each of the scenarios is highlighted in bold.}
\label{tab:extreme-scenario}
\centering
\resizebox{\linewidth}{!}{\begin{tabular}{lrrr}
\toprule
& \textbf{Label-flipping} & \textbf{Out-of-dist.} & \textbf{Random weights}\\
\toprule
\textbf{No attack} & 0,9657 & 0,9657 & 0,9657 \\
\toprule
\textbf{FedAvg} & 0,3561 & 0,4394 & 0,0994 \\
\midrule
\textbf{Trimmed-mean} & 0,6256 & 0,5778 & 0,1002 \\
\textbf{Median} & 0,8595 & 0,8347 & 0,9355 \\
\midrule
\textbf{Krum} & 0,8801 & 0,8678 & 0,8633 \\
\textbf{MultiKrum (5)} & 0,9336 & 0,9366 & 0,9349 \\
\textbf{MultiKrum (20)} & 0,9623 & 0,9617 & 0,8595 \\
\textbf{MultiKrum (25)} & 0,9623 & 0,9617 & 0,8595 \\
\midrule
\textbf{Bulyan (f=1)} & 0,4755 & 0,5005 & 0,1000 \\
\textbf{Bulyan (f=5)} & 0,9485 & 0,9475 & 0,9455 \\
\midrule
\textbf{DDaBA} & 0,4235 & 0,4819 & 0,0997\\
\textbf{SDaBA (1/4)} &  \textbf{0,9654} & \textbf{0,9653} & \textbf{0,9629}\\
\bottomrule
\end{tabular}}
\end{table}

The results show how this extreme scenario highly affects to DDaBA, but also Bulyan (f=1). With respect to the baselines, in this case it is MultiKrum with $d=20$ that achieves the best results by setting the $d$ parameter to its optimal value. Finally, we highlight the appropriate performance of SDaBA, outperforming the rest of the defenses and solving the problem of extreme scenarios.

\section{Conclusion and future work} \label{s_conclusions}

We addressed the problem of defending against byzantine attacks in FL, which is a real challenge since the existing defenses are not enough. Using the exponential distribution resulting of the differences between the best model and the rest of them in terms of accuracy over a central validation set, \corr{we consider that  those clients that represent outliers in that distribution are likely to be adversarial ones. Hence,} we propose DDaBA, a defense against byzantine attacks which dynamically filters out the adversarial and poor clients.

We \emc{evaluated the} DDaBA in three different byzantine attacks, in three datasets and using three different settings. In addition, we proposed a static version of the defense approach in order to use it in scenarios with an extremely high proportion of adversarial clients. \corr{Both the experiments corroborate the following conclusions}:
\begin{itemize}
    \item DDaBA is a highly effective defense against byzantine attacks in real attack scenarios.
    \item It properly filters out adversarial and poor clients improving the performance of the global model in scenarios with adversarial clients, even outperforming the performance in the original task.
    \item The static version SDaBA is an effective solution for extreme attack scenarios.
\end{itemize}

To conclude, we have proven that DDaBA is a high quality defense against byzantine attacks, and it can act as a proper federated aggregation operator, since it defends the global model against the effect of the attacks while improving the learning of the global model.

\correcciones{
\appendix
\section{Ordered weighted model averaging} \label{appendix:iowa}

Group decision making is the AI task focused on finding out a consensus decision from a set of experts by summing up their individual evaluations. Yager proposed in \cite{owa_1} the Ordered Weighted Averaging (OWA) operators with the aim of modelling the fuzzy opinion majority \cite{fuzzy_majority} in group decision making. Yager and Filev generalised the OWA operator definition  in \cite{iowa_yager_filev}, where they defined the OWA operator with an order-induced vector for ordering the argument variable. They called this generalisation of OWA operators with a specific semantic in the aggregation process as Induced Ordered Weighted Averaging (IOWA). The OWA and IOWA operators are weighted aggregation functions that are mathematically defined as what follows:

\begin{definition}[OWA Operator \cite{owa_1}]
An OWA operator of dimension $n$ is a function $\Phi:\mathbb{R}^n \rightarrow \mathbb{R}$ that has an associated set of weights or  weighting vector $W = (w_1, \dots, w_n)$ so that $w_i \in [0,1]$ and $\sum_{i=1}^{n}w_i = 1$, and it is defined to aggregate a list of real values $\{c_1, \dots, c_n\}$ according to the Equation \ref{eq_owa_op_apend}:

\begin{equation}\label{eq_owa_op_apend}
    \Phi(c_1, \dots, c_n) = \sum_{i=1}^{n} w_i c_{\sigma(i)}
\end{equation}
being $\sigma: \{1, \dots, n\} \rightarrow \{1, \dots, n\}$ a permutation function such that $c_{\sigma(i)} \ge c_{\sigma(i+1)}$, $\forall i = \{1, \dots, n-1\}$.
\end{definition}

\begin{definition}[IOWA Operator \cite{iowa_yager_filev}]
An IOWA operator of dimension $n$ is a mapping $\Psi: (\mathbb{R} \times \mathbb{R})^n \rightarrow \mathbb{R}$ which has an associated set of weights $W = (w_1, \dots, w_n)$ so that  $w_i \in [0,1]$ and $\sum_{i=1}^{n}w_i = 1$, and it is defined to aggregate the second arguments of a 2-tuple list  $\{\langle u_1,c_1 \rangle, \dots, \langle u_n,c_n \rangle\}$ according to the following expression:

\begin{equation}
    \Psi(\langle u_1,c_1 \rangle, \dots, \langle u_n,c_n \rangle) = \sum_{i=1}^{n} w_i c_{\sigma(i)}
\end{equation}
being $\sigma: \{1, \dots, n\} \rightarrow \{1, \dots, n\}$ a permutation function such that $u_{\sigma(i)} \ge u_{\sigma(i+1)}$, $\forall i = \{1, \dots, n-1\}$. The vector of values $U = (u_1, \dots, u_n)$ is called the order-inducing vector and $(c_1, \dots, c_n)$ the values of the argument variable. 

\end{definition}

The OWA and IOWA operators are functions for weighting the contribution of experts for the global decision in the case of group decision making, and the contribution of a set of clients in an aggregation process in a general scenario. However, they need an additional function to calculate the values of the parameters, which in the context of group decision making means the grade of membership to a fuzzy concept. The weight value calculation function is known as linguistic quantifier \cite{quantifiers}, which is defined as a function $Q:[0,1] \rightarrow [0,1]$ such as $Q(0) = 0$, $Q(1) = 1$ and $Q(x) \ge Q(y)$ for $x > y$. Equation \ref{eq_q_weight_calculation} defines how the function $Q$ computes the weight values and Equation \ref{eq_standard_quantifier_appendix} defines the behaviour of the function $Q$. 

\begin{equation}
    w_i^{(a,b)} = Q_{a,b} \left( \frac{i}{n} \right) - Q_{a,b} \left( \frac{i-1}{n} \right)
    \label{eq_q_weight_calculation}
\end{equation}

\begin{equation}\label{eq_standard_quantifier_appendix}
    Q_{a,b}(x) = 
    \begin{dcases}
        0 & 0\leq x \leq a\\
        \frac{x-a}{b-a} & a \leq x \leq b \\
        1 & b \leq x \leq 1
    \end{dcases}
\end{equation}
where $a, b \in [0,1]$ satisfying $0 \leq a \leq b \leq 1$.

The function $Q$ in Equation \ref{eq_standard_quantifier_appendix} can be redefined in order to model different linguistic quantifiers. Since the definition of the notion quantifier guided aggregation \cite{owa_1,quantifiers}, other definitions of the function $Q$ has been proposed to model different linguistic quantifiers like ``most'' or ``at least'' \cite{fuzzy_majority}.}

\section*{Acknowledgements}


This research work is supported by the R\&D\&I grants PID\-2020-119478GB-I00, PID2020-116118GA-I00 and EQC2018-005\-084-P funded by MCIN/AEI/10.13039/501100011033 and by “ERDF A way of making Europe”. Nuria Rodríguez Barroso were supported by the grant FPU18/04475 funded by MCIN\-/AEI/10.13039/501100011033 and by ``ERDF A way of making Europe''. Eugenio Martínez Cámara were supported by the grant IJC2018-036092-I funded by MCIN/AEI/10.13039/5011\-00011033.

\bibliography{mybibfile}

\end{document}